\documentclass[11pt, copyright, gdm]{google}

\usepackage[authoryear, sort&compress, round]{natbib}
\usepackage{caption}
\usepackage{lipsum}
\usepackage{listings}
\usepackage{hyperref}       
\usepackage{url}            
\usepackage{booktabs}       
\usepackage{amsfonts}       
\usepackage{nicefrac}       

\usepackage{microtype}      
\usepackage{subcaption}

\usepackage{xcolor}

\usepackage{algorithm}
\usepackage[noend]{algpseudocode}
\usepackage{mathrsfs}
\usepackage{dsfont}
\usepackage{array}
\usepackage{bbm}
\usepackage[mathscr]{eucal}
\usepackage{psfrag}
\usepackage{here}
\usepackage{wasysym}
\usepackage{amsthm}
\usepackage{todonotes}
\usepackage{tabulary}
\usepackage{outlines}
\usepackage{thmtools, thm-restate}

\newcommand{\go}[1]{\href{http://go/#1}{go/#1}}

\definecolor{ian_highlight}{RGB}{100, 2, 2}

\errorcontextlines\maxdimen

\makeatletter
    \newcommand*{\algrule}[1][\algorithmicindent]{\makebox[#1][l]{\hspace*{.5em}\thealgruleextra\vrule height \thealgruleheight depth \thealgruledepth}}%
\newcommand*{\thealgruleextra}{}
\newcommand*{\thealgruleheight}{.75\baselineskip}
\newcommand*{\thealgruledepth}{.25\baselineskip}

\newcount\ALG@printindent@tempcnta
\def\ALG@printindent{%
    \ifnum \theALG@nested>0
        \ifx\ALG@text\ALG@x@notext
        \else
            \unskip
            \addvspace{-1pt}
            \ALG@printindent@tempcnta=1
            \loop
                \algrule[\csname ALG@ind@\the\ALG@printindent@tempcnta\endcsname]%
                \advance \ALG@printindent@tempcnta 1
            \ifnum \ALG@printindent@tempcnta<\numexpr\theALG@nested+1\relax
            \repeat
        \fi
    \fi
    }%
\usepackage{etoolbox}
\makeatother

\newbox\statebox
\newcommand{\myState}[1]{%
    \setbox\statebox=\vbox{#1}%
    \edef\thealgruleheight{\dimexpr \the\ht\statebox+1pt\relax}%
    \edef\thealgruledepth{\dimexpr \the\dp\statebox+1pt\relax}%
    \ifdim\thealgruleheight<.75\baselineskip
        \def\thealgruleheight{\dimexpr .75\baselineskip+1pt\relax}%
    \fi
    \ifdim\thealgruledepth<.25\baselineskip
        \def\thealgruledepth{\dimexpr .25\baselineskip+1pt\relax}%
    \fi
    \State #1%
    \def\thealgruleheight{\dimexpr .75\baselineskip+1pt\relax}%
    \def\thealgruledepth{\dimexpr .25\baselineskip+1pt\relax}%
}

\uselogo{} 

\title{Efficient Exploration at Scale}

\correspondingauthor{vikranthd@google.com}

\reportnumber{0001} 

\author{Seyed Mohammad Asghari*, Chris Chute*, Vikranth Dwaracherla*, Xiuyuan Lu*, Mehdi Jafarnia, Victor Minden, Zheng Wen, Benjamin Van Roy \\ $ $ \\ {\bf The Efficient Agent Team} \\ {\bf Google DeepMind}}

\begin{abstract}
We develop an online learning algorithm that dramatically improves the data efficiency of reinforcement learning from human feedback (RLHF).  Our algorithm incrementally updates reward and language models as choice data is received.  The reward model is fit to the choice data, while the language model is updated by a variation of \texttt{reinforce}, with reinforcement signals provided by the reward model.  Several features enable the efficiency gains: a small affirmative nudge added to each reinforcement signal, an epistemic neural network that models reward uncertainty, and information-directed exploration.  With Gemma large language models (LLMs), our algorithm matches the performance of offline RLHF trained on 200K labels using fewer than 20K labels, representing more than a 10x gain in data efficiency.  Extrapolating from our results, we expect our algorithm trained on 1M labels to match offline RLHF trained on 1B labels.  This represents a 1,000x gain. To our knowledge, these are the first results to demonstrate that such large improvements are possible.
\end{abstract}

\begin{document}

\maketitle

\section{Introduction}
\label{sec:introduction}

While today’s large models have learned from vast amounts of data, one critical challenge going forward is to gather the right data.  Gathering more informative data can greatly accelerate learning not only of new capabilities but also of human preferences needed to guide how those capabilities are applied.  Indeed, efficient exploration should serve as a cornerstone on the path to safe artificial superintelligence.

This paper develops an algorithm for reinforcement learning from human feedback (RLHF).  Our algorithm incrementally updates reward and language models as human choices between alternative responses are observed.  The reward model (RM) is fit to the choice data, while the language model (LM) is updated by a variation of \texttt{reinforce}, with reinforcement signals provided by the RM.  Three notable innovations enable large gains in data efficiency: a small affirmative nudge added to each reinforcement signal, an epistemic neural network that models reward uncertainty, and information-directed exploration. 

\begin{figure}[ht!]
\centering
\includegraphics[width=0.7\textwidth]{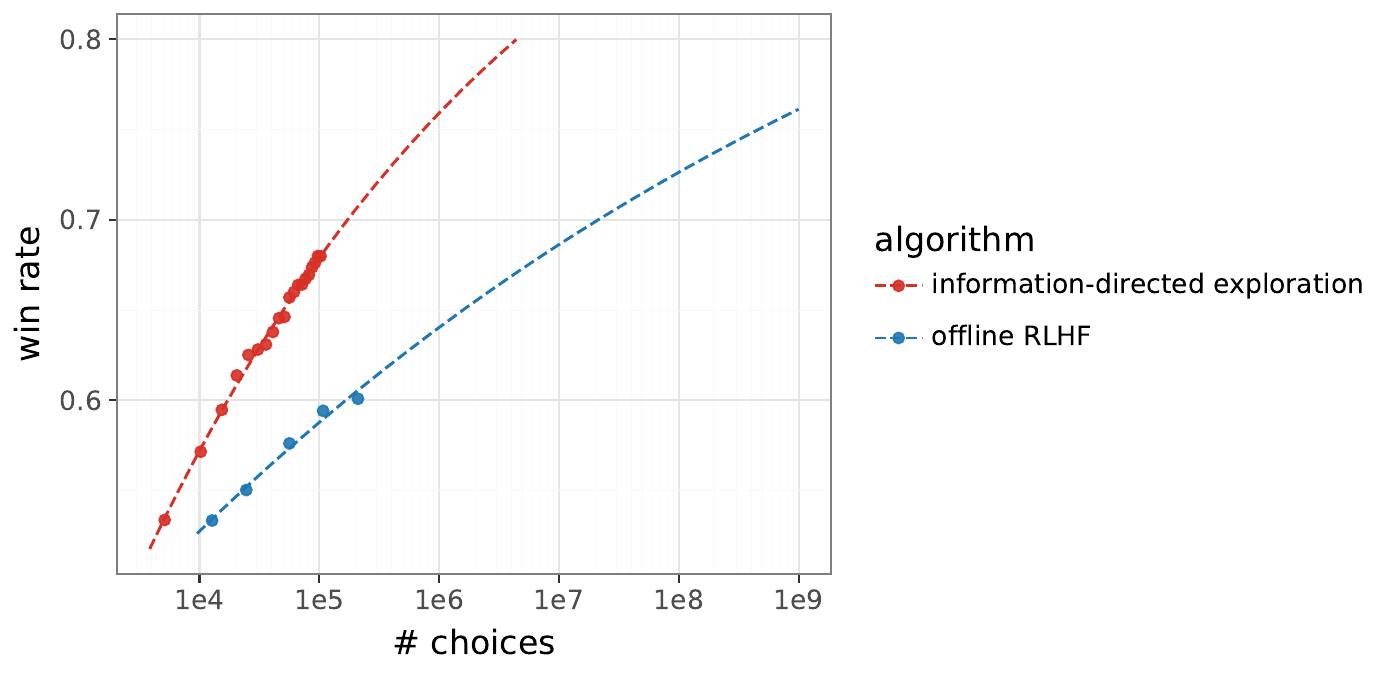}
\caption{The plots are of performance, in terms of the win rate over a baseline policy, as functions of the amount of human feedback, in terms of the number of choices observed.  Efficient exploration shifts the scaling law.}
\label{fig:extrapolation}
\end{figure}

Figure \ref{fig:extrapolation} illustrates these data efficiency gains relative to an offline RLHF baseline.  After observing fewer than 20K choices, our algorithm matches the performance of offline RLHF trained on 200K choices, representing more than a 10x gain in data efficiency. The gain is projected to grow with more choice data.  With 1M choices, our algorithm is projected to reach a gain of 1,000x.  To the best of our knowledge, these are the first results to demonstrate that such large gains are possible with large language models (LLMs).

The remainder of this report details these findings. Section~\ref{sec:literature_review} discusses relevant literature. We describe our experiment pipeline in Section~\ref{sec:experiment-pipeline} and learning algorithms in Section~\ref{sec:algorithms}. Section~\ref{sec:results} presents our empirical results, which demonstrate the data efficiency gains. Finally, we conclude and discuss promising avenues for future work in Section~\ref{sec:conclusions-and-future-work}.


\section{Literature Review}
\label{sec:literature_review}

Our work focuses on sample-efficient alignment of LLMs through active exploration and is related to work on online adaptation, active exploration, and scaling laws.

\textbf{Online Adaptation.} Work on online adaptation emphasizes iterative and sequential learning \citep{mehta2025sample, dong2024rlhf, bai2022rlhf}.  Iterative versions of direct preference optimization (DPO), as investigated by \citet{xiong2024iterativepreferencelearninghuman}, and hybrid preference optimization (HPO) \citep{bose2025hybrid} serve as examples. Recent studies demonstrate a clear advantage of online methods over their offline counterparts \citep{tang2024understanding}. This efficiency gain stems from the key difference that online algorithms sample responses on-policy, allowing them to constantly shift the response distribution toward better responses.  Offline algorithms, which use a fixed sampling distribution, suffer from challenges related to data coverage and stationary learning targets.

\textbf{Active Exploration.} Traditional work on active exploration seeks to reduce annotation costs by selecting informative examples, often relying on uncertainty or diversity metrics \citep{settles2009active}.
In LLM alignment, active exploration strategies that explicitly incorporate uncertainty and informativeness are central to improving sample efficiency. \citet{dwaracherla2024efficient}, \citet{mehta2025sample}, and \citet{ji2025reinforcement} formalize this problem as an active contextual dueling bandit.  \cite{dwaracherla2024efficient} found that uncertainty-guided exploration can significantly improve reward models.  However, their work focused exclusively on updating the RM, while the LM was fixed throughout the process. Techniques like active preference optimization (APO) and its variants, apply active learning principles directly to preference-based objectives (like DPO), iteratively collecting choice data that resolve uncertainty \citep{das2025active, ji2025reinforcement, pmlr-v235-muldrew24a}.  Techniques like exploratory preference optimization (XPO) and those based on information-directed sampling (IDS) incorporate exploration bonuses to steer the policy toward sampling where the reward model's estimates are uncertain \citep{xie2025exploratory, qi2025sample, liu2024sample}. Some other approaches harness the LLM itself to guide this exploration. Some methods utilize the model's output variability such as measuring the disagreement across multiple generated responses as a proxy for uncertainty \citep{diao2023active, bayer2026activellm}.  The aforementioned literature reports gains of 2X to 5X and works with a much more limited range of prompts than our diverse set.

\textbf{Scaling Laws.} The scaling properties of large language models, which yield considerable performance improvements with more data, are key to their success. While these scaling laws have been studied extensively for both the pre-training stage \citep{kaplan2020scaling, hoffmann2022scalinglaws} and supervised fine-tuning \citep{yuan2023scaling-sft}, how reinforcement learning from human feedback (RLHF) improves with data is less understood. Recent studies have begun to explore related aspects, such as scaling laws for reward models \citep{gao2022scalinglawsrewardmodel,rafailov2024scaling,cobbe2021trainingverifierssolvemath}. Nevertheless, a systematic understanding of how overall RLHF performance scales with the quantity of preference data remains elusive. This issue is compounded by the recent study suggesting that, disappointingly, current RLHF techniques demonstrate limited scalability, yielding only insignificant performance gains even when the quantity of preference data is substantially increased \citep{hou2024doesrlhfscaleexploring}.
This raises a critical question: can current RLHF techniques improve performance as more data becomes available or must we develop new techniques with better scaling properties?  Our work is the first to systematically study the scaling laws governing performance as a function of the amount of human data. 
 Papers proposing new RLHF algorithms often plot performance against the quantity of human data. However, plots with a linearly scaled horizontal axis obscure salient patterns.  Using instead a logarithmic scale, as in Figure \ref{fig:extrapolation}, reveals qualitative differences between scaling laws.

\section{Experiment Pipeline}
\label{sec:experiment-pipeline}

In this section, we present the experimentation pipeline used to train and compare LLM policies.  We use this pipeline to assess the performance of RLHF algorithms.

\subsection{Baseline and Experimentation Policies}

To produce a baseline policy, we use a 9B Gemma model \citep{gemmateam2024gemma2improvingopen}.  Given approximately nine billion model parameters $\theta_0$, a prompt $X$, and a partial response $Y_1,\ldots,Y_{\ell-1}$, the model specifies a next token distribution $\pi_{\theta_0}(\cdot|X, Y_1,\ldots,Y_{\ell-1})$.  The model can be used to sample a response $Y$ sequentially, token by token, until one indicates termination.

We interpret the manner in which $\pi_\theta$ samples each next token as a {\it policy}.  But we consider a class of policies, which we refer to as top-$K$ policies.  Given $X, Y_1, \ldots, Y_{\ell-1}$, the top-$K$ policy first identifies the set $\mathcal{Y}_\ell$ of $K$ tokens $y \in \mathcal{Y}_\ell$ that attain the largest probabilities $\pi_{\theta_0}(y|X, Y_1, \ldots, Y_{\ell-1})$ and then samples an element of $\mathcal{Y}_\ell$ according to the corresponding conditional probabilities.  If $K$ is equal to the number of tokens, the top-$K$ policy simply samples from the predictive distribution $\pi_{\theta_0}(\cdot|X, Y_1,\ldots,Y_\ell)$.

As a baseline, we use the top-$1$ policy with parameters $\theta_0$, which result from pretraining and supervised fine-tuning (SFT) of the Gemma model, but not RLHF.  This policy serves as a benchmark for comparison against improved top-$1$ policies based on different parameters.  Note that the baseline policy is deterministic in the sense that $X, Y_1, \ldots, Y_\ell$ determine $Y_{\ell+1}$.

To sample candidate responses for experimentation when querying human feedback, we will use top-$5$ policies.  Note that top-$5$ policies are typically stochastic.  Use of such a policy diversifies responses so that a human choice between alternative responses is informative.

\subsection{Human Feedback Simulator}

We simulate human feedback using a reward model that is based on the Gemini 1.5 Pro LLM \citep{geminiteam2024gemini15unlockingmultimodal}, trained on real human feedback.  Given a prompt $X$ and two responses $(Y_1,Y_2)$, this simulator computes two reward values $(R_1,R_2)$ and maps these to a preference probability $P = \exp(R_1) / (\exp(R_1) + \exp(R_2))$ via the Bradley-Terry model \citep{Bradley1952Rank} with an exponential score function.  Then, a simulated human choice $C \sim \mathtt{Bernoulli}(P)$ is sampled.

It is worth noting that the Gemini Pro model is much larger than 9B Gemma models.  As such, the simulated choices reflect behaviors far more complex than baseline or competing policies that we consider.  We train and test policies on such complex choice behaviors so that our results are more likely to carry over to real human choices, which may also reflect behaviors more complex than LLMs.

\subsection{Prompts}

We use a set of 202K prompts sampled from an internal repository routinely used for post-training.  These prompts cover a wide range of topics such as writing, coding, summarization, reading comprehension, math, science, ideation, etc.  Each prompt is unique and the set is randomly ordered.  Within our experiment pipeline, we use 200K prompts for training, 1K for testing and hyperparameter selection, and 1K for out-of-sample evaluation.

\subsection{Gathering Feedback and Training}

\begin{figure}[ht!]
    \centering
    \includegraphics[width=0.75\textwidth]{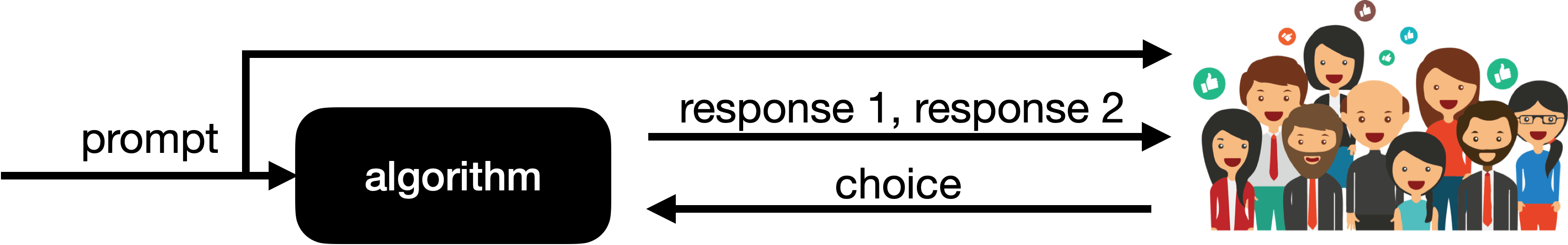}
    \caption{Feedback in the form of a choice between two responses produced is used to improve the policy.}
    \label{fig:learning_query}
\end{figure}

When gathering feedback for training, an algorithm iterates through the training prompts.  For each prompt, the algorithm generates two responses and receives a choice from the human feedback simulator, as illustrated in Figure \ref{fig:learning_query}.  Prompts are grouped into batches of 64.  After generating responses and observing choices for each batch of 64, the algorithm can adjust parameters to improve its policy based on the feedback.  We denote by $\theta_t$ the policy parameters obtained after gathering $t$ batches of choice data.

\subsection{Performance Evaluation}

\begin{figure}[ht!]
    \centering
    \includegraphics[width=0.8\textwidth]{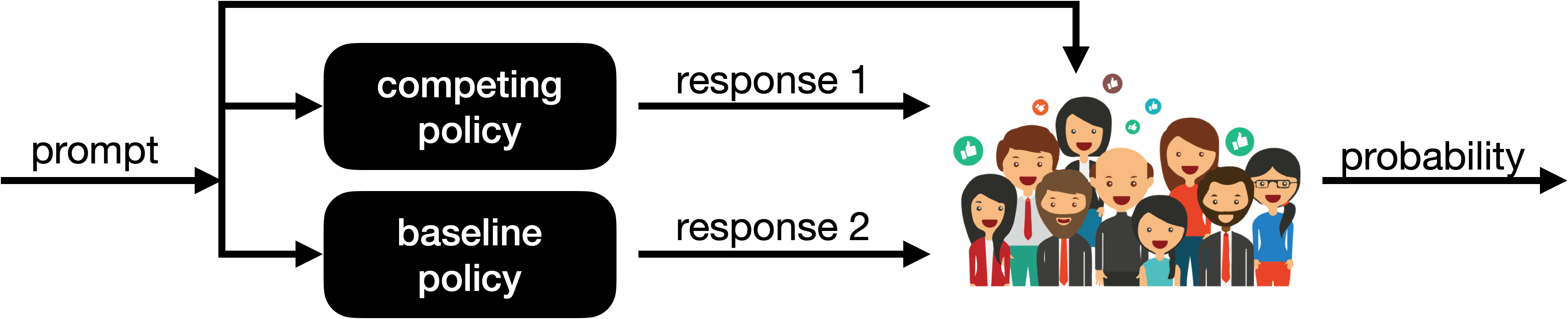}
    \caption{Given responses generated by competing and baseline policies, the human feedback simulator produces a preference probability.}
    \label{fig:eval_query}
\end{figure}

Given policy parameters $\theta_t$, we compare performance against the baseline parameters $\theta_0$ by evaluating a win rate.  This involves iterating over the 1K out-of-sample prompts.  For each such prompt $X$, we generate responses $Y_1$ and $Y_2$ using top-$1$ policies with parameters $\theta_t$ and $\theta_0$, respectively.  As illustrated in Figure \ref{fig:eval_query}, given the prompt $X$ and responses $(Y_1,Y_2)$, the human feedback simulator produces a preference probability $P$.  The win rate $\overline{P}$ is an average of the preference probability over the 1K out-of-sample prompts.  An outcome of $\overline{P} = 1$ implies that the competing model is always chosen over the baseline, while an outcome of $\overline{P} = 0$ implies the opposite.  Intermediate values of $\overline{P}$ indicate decrease of preference for the competing model over the baseline.

\section{Algorithms}
\label{sec:algorithms}

We will compare performance across four alternative algorithms.  While we will later explain these alternatives in greater detail, the following brief descriptions differentiate each:
\begin{enumerate}
\item {\bf offline RLHF:} Gather $T$ batches of choice data with responses generated using $\pi_{\theta_0}$, then fit a reward model (initialized with $\phi_0$), then optimize the policy (initialized with $\theta_0$) to produce $\theta_T$.
\item {\bf periodic RLHF:} For some period $\tau$ that is a fraction of $T$, produce a sequence of policies $\theta_{k \tau}: k = 1, 2,\ldots, T/\tau$.  For each $k$, gather $\tau$ batches with responses generated by $\pi_{\theta_{(k-1)\tau}}$, then fit a reward model (to $k\tau$ batches, initialized with $\phi_0$), then optimize the policy (initialized with $\theta_0$) to produce $\theta_{k\tau}$.
\item {\bf online RLHF:} Generate a sequence of policies $\pi_{\theta_t}: t = 1, 2,3,\ldots$.  For each $t$, gather a batch with responses generated using $\pi_{\theta_{t-1}}$, then incrementally adjust the reward model, then incrementally adjust $\theta_{t-1}$ to produce $\theta_t$.
\item {\bf information-directed exploration:} Apply online RLHF but incrementally adjust a model of reward uncertainty alongside the point estimate reward model and use that to guide response selection.
\end{enumerate}
In each case, $\phi_0$ and $\theta_0$ denote initial parameters of the reward model and policy.  These four alternatives were developed through trial and error, in each case iterating over algorithm designs and hyperparameters while adhering to the above descriptions.  Reward models and policies across these alternatives use approximately the same number of parameters.  Also, parameter update rules are similar across the alternatives.  We will describe common elements across reward models, policies, and update rules, and then explain each alternative in greater detail.  Our intention is not to provide sufficient detail to reproduce our results but to share salient elements of our algorithms along with some motivation.

\subsection{Reward Models and Policies}

The initial parameters $\theta_0$ result from pretraining and supervised fine-tuning (SFT) of a Gemma 9B model.  Each reward model is initialized with the same transformer backbone -- that is, the the same language model $\pi$ with the unembedding matrix and softmax removed.  The output of the backbone, which we refer to as the {\it last-layer embedding}, is then mapped to a scalar reward via a {\it head}, initialized with random weights.

For offline RLHF, periodic RLHF, and online RLHF, we use a linear head.  For information-directed sampling, we use an ensemble of multilayer perceptron (MLP) heads, as we will motivate and describe further later.  Use of an MLP rather than linear head did not improve the performance of other algorithms.

We denote the initial reward model by $r_{\phi_0}$, where $\phi_0$ is the vector of parameters, including those of the backbone and the head.  Each of our algorithms update parameters of reward model and policy.  

\subsection{Update Rules}

We now discuss update rules that we use to adjust reward model and policy parameters.  Given a prompt $X$ and two responses $Y$ and $Y'$ a reward model $r_{\phi_t}$ predicts the probability that $Y$ will be chosen over $Y'$:
\begin{equation}
p_{\phi_t}(Y \succeq Y'|X) = \frac{e^{r_{\phi_t}(Y|X)}}{e^{r_{\phi_t}(Y|X)} + e^{r_{\phi_t}(Y'|X)}}.
\end{equation}
Suppose a prompt and pair of responses is presented to a rater, and we denote the chosen response by $Y$ and the other by $Y'$.  Then, our reward model update rule computes a gradient
\begin{equation}
\label{eq:reward-gradient}
\Delta \phi_t = \nabla_{\phi_t} \ln p_{\phi_t}(Y \succeq Y'| X).
\end{equation}
These gradients are summed over a batch of prompts.  We will later explain how each of our algorithms selects a pair of responses for each prompt. The gradients are summed over a batch of prompts, clipped, and then used to update $\phi_t$ and obtain $\phi_{t+1}$ using AdamW  \citep{Loshchilov2017adamw}.

The policy update rule is slightly more complex.  It entails maintaining an exponential moving average of parameters
\begin{equation}
\overline{\theta}_{t+1} = \eta \overline{\theta}_t + (1-\eta) \theta_{t+1},
\end{equation}
for some $\eta \in (0,1)$.  We refer to $\overline{\theta}_t$ as an anchor.  Parameters $\theta_t$ are regularized toward this anchor as they are updated.  Given a prompt $X$ and a pair $(Y,Y')$ of responses, our update rule computes a policy gradient
\begin{equation}
\label{eq:policy-gradient}
\Delta \theta_t = \left(p_{\overline{\phi}_t}(Y \succeq Y' | X) - \frac{1}{2}\right) \nabla_{\theta_t} \ln \pi_{\theta_t}(Y|X) - \beta \sum_{\ell=1}^{\mathrm{len}(Y)} \nabla_{\theta_t} \mathrm{KL}\left(\pi_{\overline{\theta}_t}(\cdot | X, Y_{1:\ell-1}) \| \pi_{\theta_t}(\cdot | X, Y_{1:\ell-1})\right).
\end{equation}
The parameter $\beta$ weights the degree of regularization toward the anchor.  Note that $\overline{\phi}_t$ denotes reward function weights used to update $\theta_t$; our online algorithms use $\overline{\phi}_t = \phi_t$, while other algorithms use different parameters.  The update rule can be viewed as a variant of \texttt{reinforce} \citep{sutton2018reinforcement} with reinforcement signal $p_{\phi_t}(Y \succeq Y' | X) - 1/2$ or, alternatively, as a variant of \texttt{PMPO} \citep{abdolmaleki2025learningnegativefeedbackpositive}.

Policy gradients are summed over a batch of prompts and multiple response pairs assigned to each prompt.  We will later explain how each of our algorithms selects these response pairs.  Policy gradients are summed, clipped, and then used to update policy parameters using AdamW.  The gradients are summed over a batch of prompts, clipped, and then used to update using AdamW  \citep{Loshchilov2017adamw}.  One or more such adjustments are made to generate the difference $\theta_{t+1} - \theta_t$ between gathering the $t$th and $(t+1)$th batches of choice data.

\subsection{Alternatives}

We now explain in greater detail key features of each alternative algorithm beyond the aforementioned common elements.

\subsubsection{Offline RLHF}

Recall that our offline RLHF algorithm gathers $T$ batches of choice data, with response pairs sampled independently using top-$5$ sampling with parameters $\theta_0$.  For each $t$-th batch, the reward model update rule (\ref{eq:reward-gradient}) is applied to adjust parameters, taking them from $\phi_{t-1}$ to $\phi_t$.

In order to optimize the policy, we assign to each prompt two (ordered) response pairs: a random pair sampled by the top-$5$ policy with parameters $\theta_t$ and those two responses in reverse order.  Given a batch of prompts with two response pairs assigned to each, the policy update rule (\ref{eq:policy-gradient}) with $\overline{\phi}_t = \phi_T$ is applied, with all gradients summed and then clipped to adjust parameters, taking them from $\theta_t$ to $\theta_{t+1}$.  Every 160th parameter vector $\theta_0, \theta_{160}, \theta_{320}, \ldots, \theta_T$ is stored.  The final policy parameters are taken to be the vector among these checkpoints that maximizes win-rate on the test set.

\subsubsection{Periodic RLHF}

Periodic RLHF operates much in the same way as offline RLHF.  Given a period $\tau$ that is a fraction of $T$, periodic RLHF carries out offline RLHF over $\tau$ instead of $T$ batches.  This produces policy parameters $\theta_\tau$.  Then, an additional $\tau$ batches of choice data are gathered but using response pairs sampled by the top-$5$ policy with parameters $\theta_\tau$ instead of $\pi_{\theta_0}$, a reward `function initialized at $r_{\phi_0}$ and a policy initialized at $\pi_{\theta_0}$ are updated just as in offline RLHF but using the $2\tau$ batches of data gathered so far.  This process repeats, gathering $\tau$ batches of data using each new vector of policy parameters until $T$ batches have been gathered and processed.  In our experiments, we use a period of $\tau = 400$ batches.

Relative to offline RLHF, periodically updating the language model in this way is known to improve performance \citep{bai2022rlhf}.  The degree of improvement grows as the period $\tau$ shrinks.  However, the number of times one trains new models also grows, and the approach becomes computationally onerous.  Our online algorithm overcomes this obstacle by incrementally updating the RM and LM as choice data is observed, instead of training new models and policies from scratch.

\subsubsection{Online RLHF}

\begin{figure}[h]
\centering
\includegraphics[width=0.48\textwidth]{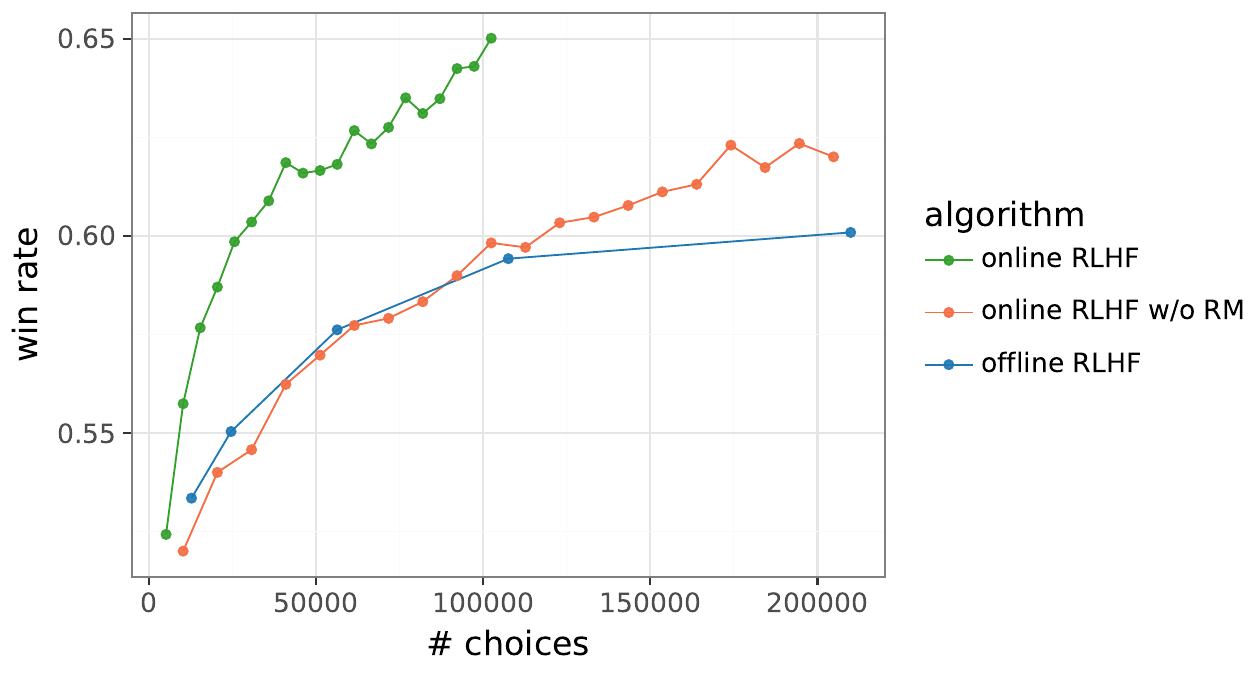} $\quad$
\includegraphics[width=0.48\textwidth]{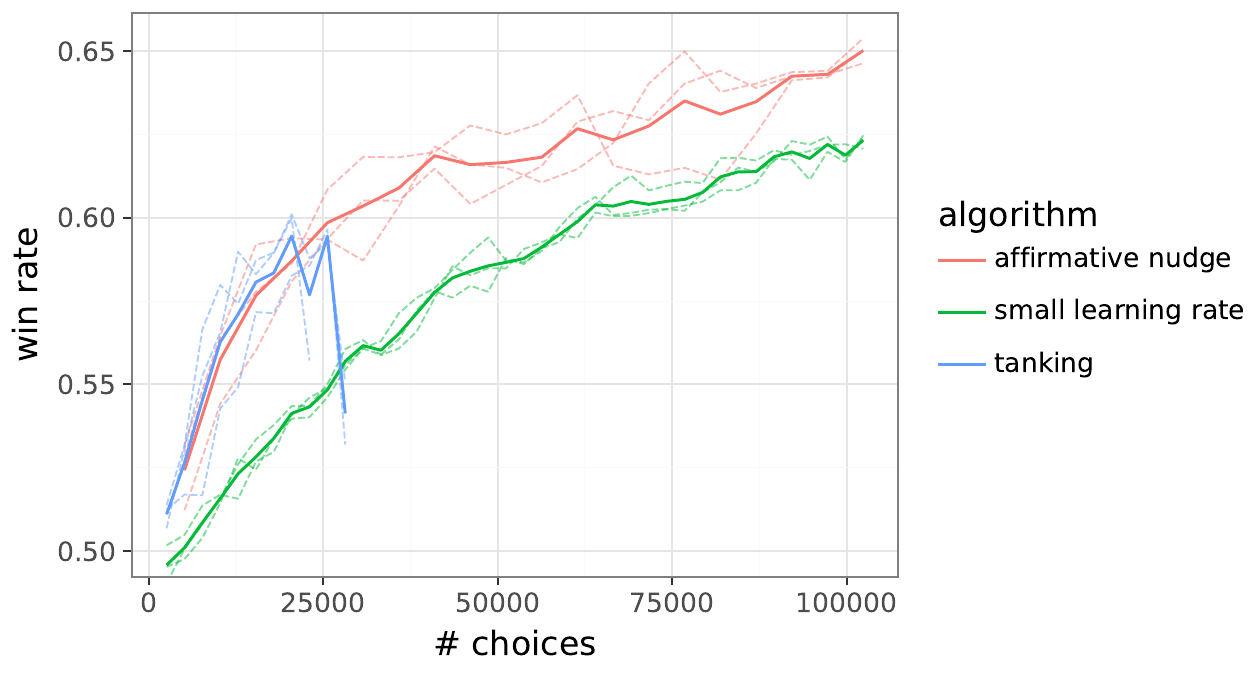}
\caption{Strong performance of online RLHF relies on a reward function (left) and an affirmative nudge (right).}
\label{fig:online_innovations}
\end{figure}

Our online RLHF algorithm interleaves between updates of reward model and policy parameters.  While prior work \citep{guo2024directlanguagemodelalignment,lin2026activedpo,belakaria2025sharpe,ji2025reinforcement} has suggested promise in updating a policy directly, without a reward model, reward-model-free approaches we have tried have not proved to be competitive.  Figure \ref{fig:online_innovations}(left) plots the best performance we were able to obtain by updating a policy without a reward model.  While showing some improvement over offline RLHF, the results are not competitive with our online RLHF algorithm, which does use a reward model.

Prior online RLHF algorithms tank after training on some number of batches, as illustrated in Figure \ref{fig:online_innovations}(right).  To address this, one can checkpoint previous policies and use one from before tanking or reduce the learning rate to delay tanking.  As illustrated in the figure, each of these solutions sacrifices performance relative to our online RLHF algorithm.  Via a slight modification of the policy update rule (\ref{eq:policy-gradient}), our algorithm avoids tanking without requiring a learning rate reduction.  This is accomplished by adding a small positive scalar $\epsilon$, which we refer to as an {\it affirmative nudge}, to each reinforcement signal.  The update rule becomes
\begin{equation}
\label{eq:affirmative-nudge}
\Delta \theta_t = \left(p_{\overline{\phi}_t}(Y \succeq Y' | X) - \frac{1}{2} + \epsilon\right) \nabla_{\theta_t} \ln \pi_{\theta_t}(Y|X) - \beta \sum_{\ell=1}^{\mathrm{len}(Y)} \nabla_{\theta_t} \mathrm{KL}\left(\pi_{\overline{\theta}_t}(\cdot | X, Y_{1:\ell-1}) \| \pi_{\theta_t}(\cdot | X, Y_{1:\ell-1})\right).
\end{equation}
Figure \ref{fig:online_innovations}(right) demonstrates the benefit.

Our algorithm updates reward model parameters as follows.  Given $\phi_t$ and $\theta_t$, for each prompt in a batch of 64, we sample sixteen responses using $\pi_{\theta_t}$ and query the human feedback simulator with a random selection of two among the sixteen.  Given choices made by the simulator, the reward model update rule (\ref{eq:reward-gradient}) is applied to adjust parameters, taking them from $\phi_t$ to $\phi_{t+1}$.

To update policy parameters, we first compute gradients according to (\ref{eq:affirmative-nudge}), with $\overline{\phi}_t = \phi_{t+1}$.  For each of the aforementioned 64 prompts, we compute gradients for four pairs of responses.  The first two are the response pair used in the query and the same pair in reverse order.  The other two are selected from the sixteen samples based on reward estimates: (highest, lowest) and (lowest, highest).  Note that rewards are assessed according to $r_{\phi_{t+1}}$.  The gradients are summed and clipped and the result added to $\theta_t$.  Then, for each of a new batch of 64 prompts, we sample 16 responses and select four pairs from among them based on reward estimates: (highest, lowest), (lowest, highest), (second-highest, second-lowest), and (second-lowest, second-highest).  We again compute gradients according to (\ref{eq:affirmative-nudge}) for each of these prompts with each of its four response pairs.  Finally, these gradients are summed, clipped, and added to parameters to produce $\theta_{t+1}$.

\subsubsection{Information-Directed Exploration}

Our information-directed sampling algorithm relies on supplementing the reward model head with components that require a very small number of additional parameters relative to the overall number, which is around nine billion.  These components enable uncertainty modeling.  We use the uncertainty estimates to guide selection of responses when constructing queries for human feedback.  The training is done in a manner similar to our online RLHF algorithm, except that we additionally train the new head components.  We now offer further detail on the architecture, queries, and training.

\subsubsection*{Architecture}

Our architecture includes a point estimate, which is identical to the architecture used by other algorithms except with the linear head replaced by an MLP with two hidden layers, each of width 1024, and a linear output node.  We call this point estimate head $\mathtt{mlp 0}$.  We supplement this head with an ensemble of MLPs, including 100 prior networks, each with two hidden layers of width 256, and 100 differential networks, each with two hidden layers of width 1024.  We call the prior networks $\mathtt{\overline{mlp}\ 1}, \ldots, \mathtt{\overline{mlp}\ 100}$ and the differential networks $\mathtt{mlp\ 1}, \ldots, \mathtt{mlp\ 100}$.  All of our MLPs are initialized with random weights.  The prior and differential networks form an ensemble with randomized prior functions, as studied in \citep{osband2018rpf}.

\begin{figure}[h]
\centering
\includegraphics[scale=0.3]{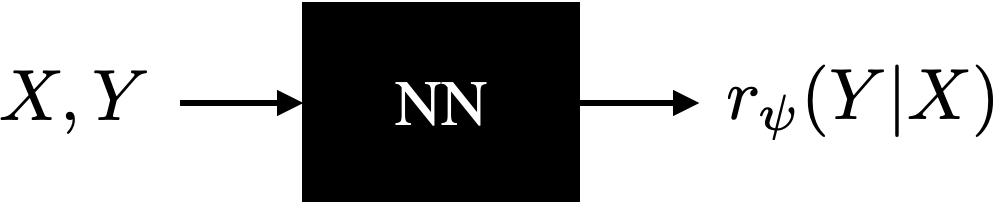} \qquad \qquad 
\includegraphics[scale=0.3]{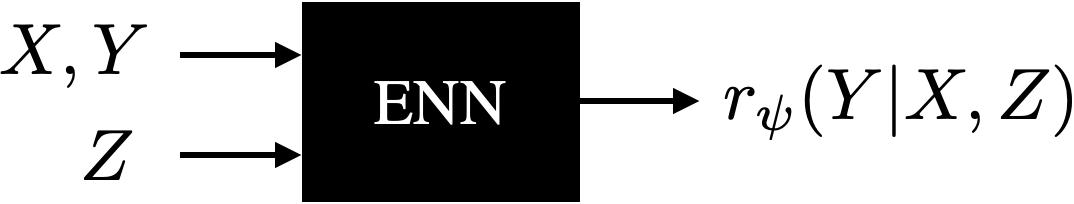}
\caption{A neural network reward model versus an epistemic neural network reward model.}
\label{fig:enn}
\end{figure}

Our architecture serves as an epistemic neural network (ENN), as studied in \citep{osband2023epistemic}.  While the reward model architecture used by previous algorithms takes as input a prompt $X$ and response $Y$ and outputs a reward $r_\phi(Y|X)$, as illustrated in Figure \ref{fig:enn}(left), the ENN reward model takes an additional input $Z$ and outputs a reward $r_\phi(Y|X, Z)$ that depends on $Z$.  We denote choice probabilities by
\begin{equation}
p_{\phi_t}(Y \succeq Y'|X,Z) = \frac{e^{r_{\phi_t}(Y|X,Z)}}{e^{r_{\phi_t}(Y|X,Z)} + e^{r_{\phi_t}(Y'|X,Z)}}.
\end{equation}
\begin{equation}
\arg\max_{Y, Y'} \rm{Var}\left[p_{\psi}(Y \succeq Y'|X,Z)\right]
\end{equation}

\begin{figure}[h]
\centering
\includegraphics[scale=0.3]{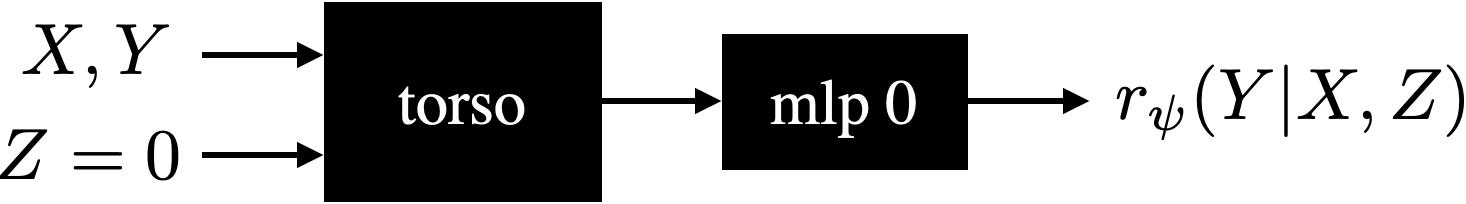}
\caption{Inference pathway for the point estimate ($Z=0$).}
\label{fig:point_estimate}
\end{figure}

\begin{figure}[h]
\centering
\includegraphics[scale=0.3]{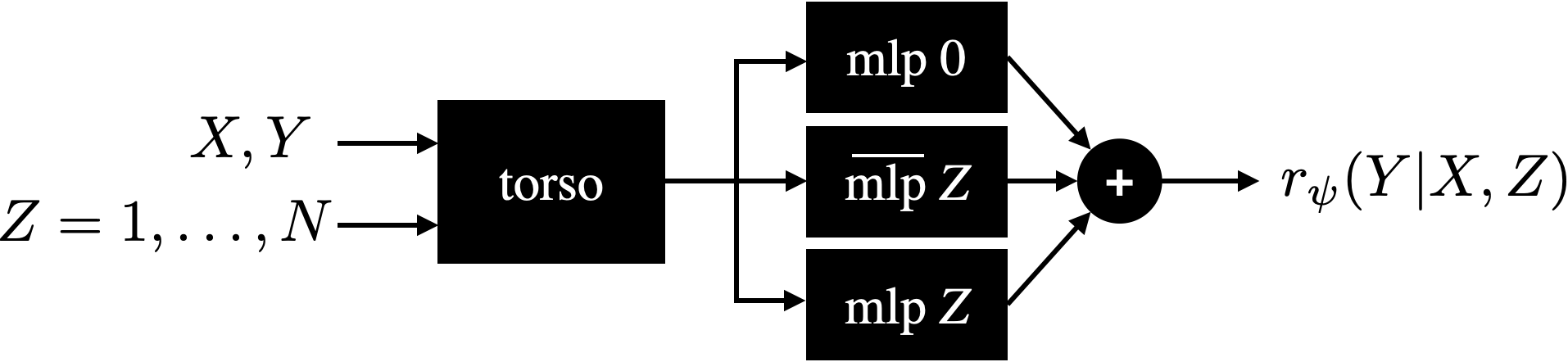}
\caption{Inference pathway for an ensemble particle ($Z = 1,\ldots,N$).}
\label{fig:ensemble}
\end{figure}

We refer to $Z$ as an {\it epistemic index}.  In our ENN reward model, $Z$ can take on integer values from $0$ to $100$.  If $Z=0$, the model carries out inference via the point estimate reward model, as illustrated in Figure \ref{fig:point_estimate}.  On the other hand, if $Z > 0$ then values of the $Z$th prior and differential networks are added, as illustrated in Figure \ref{fig:ensemble}.

While our ENN reward model requires an increase in the number of parameters, the increase is small relative to the nine billion parameters in the original model.  In particular, the percentage increase is well under 5\%.

\subsubsection*{Queries}

Given policy and reward model parameters $\phi_t$ and $\theta_t$, for each prompt in a batch of 64, we sample sixteen responses using  $\pi_{\theta_t}$.  For each prompt $X$ and pair $(Y,Y')$ of responses, we compute the variance $p_{\phi_t}(Y \succeq Y'|X,Z)$ over ensemble particles $Z=1,\ldots,100$.  We then select the response pair $(Y_*, Y_*')$ that maximizes this variance and send the query $(X,Y_*,Y_*')$ to the human feedback simulator.  The motivation is to select a response pair for which a choice will be informative, and the choice probability variance serves as a measure of informativeness.

\subsubsection*{Training}

Given a batch of 64 queries and choices, we update the point estimate reward model in exactly the same way as done in our online RLHF algorithm.  In particular, we use the same loss function but based on the output $r(Y|X,0)$, and we leave the prior and differential networks untouched in this step.  We then update each of the differential networks, individually, using the same loss function, with losses computed based on $r(Y|X,Z)$ for $Z=1,\ldots,100$.  When updated each of these differential networks, the backbone parameters are frozen.  The prior networks are never updated in the training process.

\section{Results}
\label{sec:results}

We now present and interpret results produced by applying each of the aforementioned algorithms in our experiment pipeline.

\subsection{Win Rates}

\begin{figure}[h]
\centering
\includegraphics[width=0.7\textwidth]{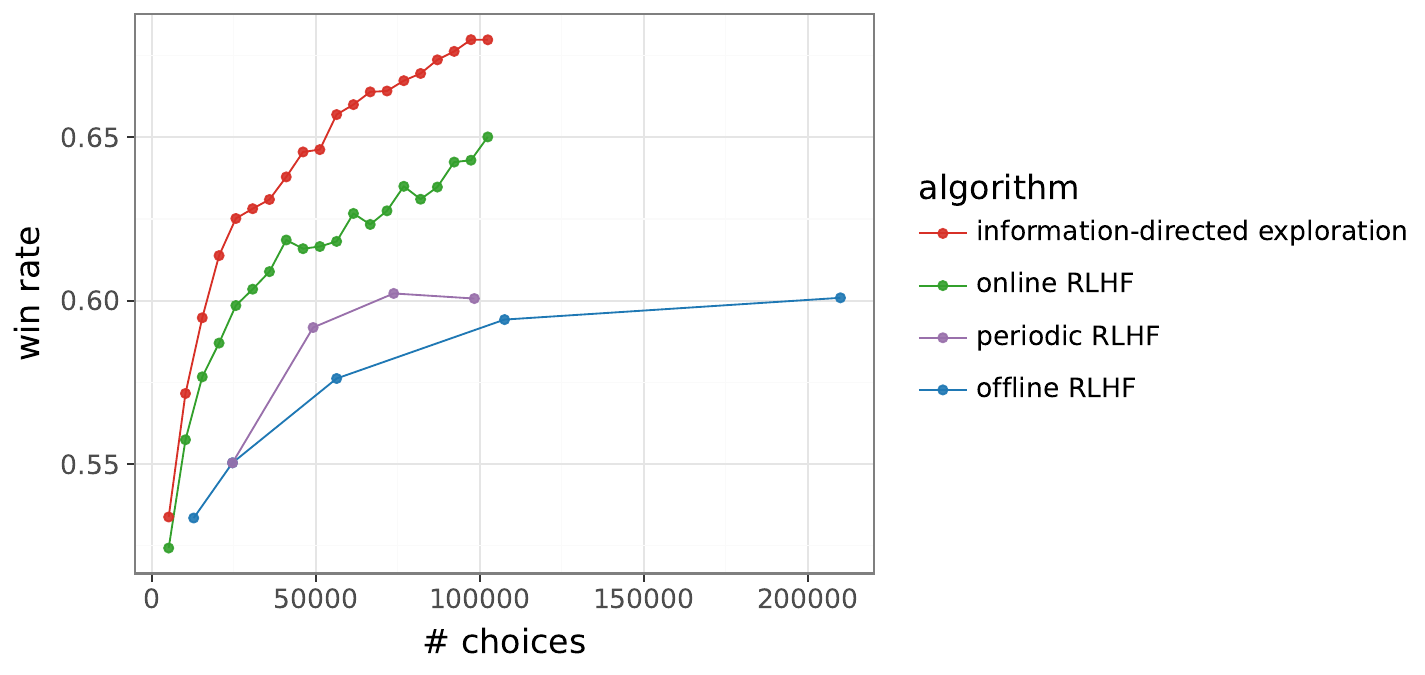}
\caption{Win rate against the baseline policy as a function of the amount of feedback collected to adapt the language model.}
\label{fig:win_rate}
\end{figure}

Figure \ref{fig:win_rate} plots win rates attained by each of our algorithms as a function of the number of choices observed.  While the win rate of offline RLHF improves with the number of choices, the alternatives do much better.  Information-directed exploration, in particular, demonstrates large improvement.  Offline RLHF needs more than 200K choices to match that performance at 20K choices.  This represents an over 10x gain in data efficiency.

\subsection{Projected Gains}

\begin{figure}[h]
\begin{subfigure}{0.6\textwidth}
\includegraphics[width=0.95\textwidth]{figures/extrapolation.pdf}
\caption{Extrapolation of win rates.}
\label{fig:extrapolation_subfigure}
\end{subfigure}
\begin{subfigure}{0.38\textwidth}
\includegraphics[width=0.95\textwidth]{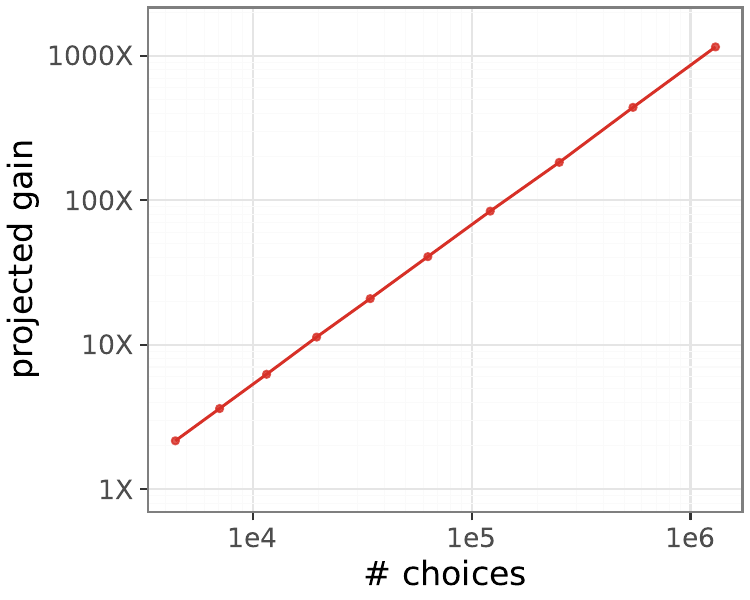}
\caption{Projected data efficiency gain.}
\label{fig:projected_gain}
\end{subfigure}

\caption{
(a) Extrapolation of win rates attained by efficient exploration and offline RLHF. We fit a function of the form $w(n) = 1 - 0.5 (n / a)^{-b}$, with parameters $a$ and $b$.
(b) Projected gain in data efficiency relative to offline RLHF. The projected gain is computed using extrapolations from (a). With 1M labels, the efficient exploration algorithm is predicted to attain a 1,000x gain in data efficiency.
}
\end{figure}

To further interpret the implications of these gains, Figure \ref{fig:extrapolation_subfigure} extrapolates win rates attained by information-directed exploration and offline RLHF, and Figure \ref{fig:projected_gain} maps these curves to projected data efficiency gains relative to offline RLHF.  These gains increase with the number of choices, leading to a gain of about 1,000x after a million choices are observed by the online algorithm.

\subsection{Simple and Extreme Instances}

Instances of prompts and responses foster intuition on what drives the data efficiency gains that we have demonstrated.  In this section, we present some that are simple so that they are easy to digest and extreme so that they emphasize salient behavior.

\subsection{Winning Responses}

Our first example illustrates how a winning response generated by information-directed exploration improves over one generated by offline RLHF.

Consider the following simple prompt, which asks for the answer to a multiple choice math problem.

\begin{small}
\colorbox{yellow}{\begin{tabular}{ll}
{\bf prompt:} & If a person walks at 14 km/hr instead of 10 km/hr, he would have walked 20 km \\
& more. The actual distance traveled by him is: \\ \\
& A. 50 km \\ & B. 56 km \\ & C. 70 km \\ & D. 80 km
\end{tabular}}
\end{small}

A model produced by the offline RLHF method generates the response below, which is convoluted and arrives at a wrong answer.

\begin{small}
\colorbox{VioletRed}{
\begin{tabular}{ll}
{\bf offline RLHF:} &
Here's how to solve the problem:\\ \\
& 1. Understand the concept of time: The key is to realize that the \\
& time taken to cover a distance remains the same in both scenarios \\
& (walking at 10 km/hr and 14 km/hr).\\ \\
& 2. Set up variables: \\ \\ 
& Let 'd' be the actual distance traveled. \\
& Let 't' be the time taken to travel the distance. \\ \\
& 3. Formulate equations: \\ \\
& We can express the distance in terms of speed and time for both \\
& scenarios: \\ \\
& $d = 10t$ (distance at 10 km/hr)\\
& $d + 20 = 14t$ (distance at 14 km/hr, which is 20 km more) \\ \\
& 4. Solve for 'd': \\ \\
& Now we have two equations, and we can solve for 'd'. One way is \\
& through substitution: \\ \\
& 1. Solve the first equation for 't': $t = d/10$ \\ \\
& 2. Substitute this value of 't' into the second equation: \\
& $d + 20 = 14(d/10)$ \\ \\
& 3. Simplify and solve for 'd': $d + 20 = (7/5)d$ \\ 
& $(3/5)d = 20$ \\
& $d = 100/3$ km \\ \\
& Therefore, the actual distance traveled is approximately 33.33 km. \\
&  None of the provided answer choices are correct.
\end{tabular}}
\end{small}

A model produced by information-directed exploration, on the other hand, generates a more concise response and, importantly, coherent logic leading up to the correct answer.

\begin{small}
\colorbox{YellowGreen}{\begin{tabular}{ll}
{\bf efficient exploration:} & Here's how to solve the problem: \\ \\
& 1. Time Difference: The person covers the extra 20 km because of \\
& the increased speed (14 km/hr - 10 km/hr = 4 km/hr). To cover \\
& 20km at this extra speed, it takes him 20 km / 4 km/hr = 5 hours. \\ \\
& 2. Distance at Original Speed: Since he walked for 5 hours at his \\
& original speed of 10 km/hr, the actual distance traveled is \\
& 5 hours * 10 km/hr = 50 km. \\ \\
& Therefore, the answer is A.
\end{tabular}}
\end{small}

\subsection{Diverse Responses}

Recall that our information-directed exploration algorithm selects responses to maximize the variance of choice probability based on an ENN reward model.  We next present two examples that illustrate how this elicits informative feedback.  For each prompt, we generated sixteen responses and select two response pairs: one that maximizes choice probability variance and one that minimizes choice probability variance.  We refer to these as {\it infomax} and {\it infomin} response pairs.

Note that whether a pair is expected to elicit informative feedback changes as labels are collected and our uncertainty model is trained on that data.  For example, if the uncertainty model infers from data collected thus far that one response is highly likely to be preferred over another then it will attribute little expected information gain to the pair.  To produce the examples below, we used the initial ENN reward model parameters $\theta_0$.

Our first example prompts for sentiment expressed by a comment on water intake.

\begin{small}
\colorbox{yellow}{\begin{tabular}{ll}
{\bf prompt:}
& Is this a negative or a positive sentiment? Do not provide reasons. \\ \\
& Keep in mind that about 20\% of our total water intake comes not from \\
& beverages but from water-rich foods like lettuce, leafy greens, cucumbers, \\
& bell peppers, summer squash, celery, berries, and melons.
\end{tabular}}
\end{small}

Responses in the infomin pair express the same content.

\begin{small}
\colorbox{VioletRed}{\begin{tabular}{ll}
{\bf infomin 1:} & Positive. \\ \\
{\bf infomin 2:} & Positive sentiment.
\end{tabular}}
\end{small}

Responses in the infomax pair carry differing meanings and therefore elicit an informative choice.

\begin{small}
\colorbox{YellowGreen}{\begin{tabular}{ll}
{\bf infomax 1:} & positive \\ \\
{\bf infomax 2:} & Neutral.
\end{tabular}}
\end{small}

Our second example entails an exercise in reading comprehension.

\begin{small}
\colorbox{yellow}{\begin{tabular}{ll}
{\bf prompt:}
& According to the article below, what was the main reason for executive order \\
& 6102? \\ \\
& [begin article] \\
& The stated reason for the order was that hard times had caused a "hoarding" \\
& of gold, stalling economic growth and worsening the depression as the US was \\
& then using the gold standard for its currency. \\ \\
& On April 6, 1933, The New York Times wrote, under the headline Hoarding of \\
& Gold, "The Executive Order issued by the President yesterday amplifies and \\
& particularizes his earlier warnings against hoarding. On March 6, taking \\
& advantage of a wartime statute that had not been repealed, he issued \\
& Presidential Proclamation 2039 that forbade the hoarding 'of gold or silver \\
& coin or bullion or currency', under penalty of \$10,000 fine or ten years \\
& imprisonment or both." \\ \\
& The main rationale behind the order was actually to remove the constraint on \\
& the Federal Reserve preventing it from increasing the money supply during \\
& the depression. The Federal Reserve Act (1913) required 40\% gold backing \\
& of Federal Reserve Notes that were issued. By the late 1920s, the Federal \\
& Reserve had almost reached the limit of allowable credit, in the form of \\
& Federal Reserve demand notes, which could be backed by the gold in its \\
& possession. \\
& [end article]
\end{tabular}}
\end{small}

As with our first example, the infomin pair presents two responses that make equivalent statements.

\begin{small}
\colorbox{VioletRed}{\begin{tabular}{ll}
{\bf infomin 1:} 
& The article states that the main rationale behind Executive Order 6102 \\ 
& was to remove the constraint on the Federal Reserve preventing it from \\
& increasing the money supply during the depression. \\ \\
{\bf infomin 2:} 
& The article states the main reason for executive order 6102 was to \\
& remove the constraint on the Federal Reserve preventing it from \\
& increasing the money supply during the depression. 
\end{tabular}}
\end{small}

The infomax pair, on the other hand, provides responses with substantially different forms of reasoning to support the conclusion.

\begin{small}
\colorbox{YellowGreen}{\begin{tabular}{ll}
{\bf infomax 1:} 
& The main reason Executive Order 6102 was issued was to remove \\ 
& the constraint on the Federal Reserve preventing it from increasing \\
& the money supply to fight the Great Depression. \\ \\
& Here's why: \\ \\
& $\bullet$ The gold standard:  The US was using the gold standard, meaning \\
& the value of the dollar was fixed to a certain amount of gold.  This \\
& system limited the Fed's ability to create new money to stimulate \\
& the economy. \\ \\
& $\bullet$ The Depression's impact: The Great Depression had severely \\
& restricted lending and economic activity. The Fed's limited ability \\
& to increase the money supply hindered efforts to boost the economy. \\ \\
& $\bullet$ Executive Proclamation 2039: This order expanded upon President \\
& Roosevelt's earlier Proclamation 2039, which prohibited gold and \\
& silver hoarding. This further aimed to reduce the supply of gold, \\
& which could have helped the Fed increase the money supply.  \\ \\
{\bf infomax 2:} 
& The main reason for Executive Order 6102, according to the \\
& provided article, was to remove the constraint on the Federal \\
& Reserve preventing it from increasing the money supply during the \\
& depression. \\ \\
& The article states: "The main rationale behind the order was \\
& actually to remove the constraint on the Federal Reserve preventing \\
& it from increasing the money supply during the depression." \\ \\
& The article also explains that the Federal Reserve Act of 1913 \\
& required 40\% gold backing for their notes, which they had almost \\
& reached by the late 1920s. This restriction prevented the Fed from \\
& creating more money to stimulate the economy. The order allowed \\
& the Fed to bypass this restriction and increase the money supply.
\end{tabular}}
\end{small}

\newpage

\section{Conclusions and Future Work}
\label{sec:conclusions-and-future-work}

We have demonstrated that online learning and uncertainty-guided exploration each offer enormous gains in data efficiency.  To realize the gains from online learning, we needed to overcome the tanking behavior commonly observed in the course of RLHF.  We did this by introducing an affirmative nudge.  Our specific approach uncertainty-guided exploration relied on training an epistemic neural network and selecting responses to maximize a measure of information gain.

While our algorithms point to a potential 1,000x data efficiency gain relative to offline RLHF, we believe there remains much room for improvement through further research.  Beyond improving algorithms, there is opportunity to translate benefits to other large language model use cases.  Some promising research directions that could build on this work include:
\begin{itemize}
\item {\bf improving the exploration algorithm.}  There are a number of ways to improve our algorithm that warrant further research.  Among them are modeling uncertainty deeper in reward models, representing uncertainty not just about the reward model but also the language model, and developing more efficient optimization algorithms that train the language model using information from the reward model. \\

\item {\bf selecting prompts.} While the exploration algorithm we presented is designed to select responses given a prompt, it can be extended to select prompts that are expected to yield informative feedback. \\

\item {\bf multiturn dialog.} Our pipeline and methods can be extended to optimize the quality of multiturn conversations.  One approach, introduced by \cite{marklund2024choice}, involves learning not only a reward model but also a value model that predicts anticipated rewards.  Our approach can be extended to work with such value models. \\

\item {\bf agents.} Our methods could also be extended to optimizing the performance of AI agents. Again, ideas from \citep{marklund2024choice} may be helpful to learning from human feedback when actions induce delayed consequences. Our approach can be extended to this work to produce better agents.\\

\item {\bf AI assisted feedback.} As AI capabilities advance and responses become more complex, comparison of responses becomes too challenging for humans. A promising direction is AI-assisted feedback, where the AI actively frames the comparison to elicit more informative feedback. For instance, a human could validate an AI-generated rationale for why one response is superior to another. The debate paradigm of \cite{irving2018ai} offers an example. Our methods could be applied to this richer feedback structure to further enhance its informativeness.
\end{itemize}

\section*{Acknowledgments}

This work benefited from stimulating conversations with many colleagues at DeepMind and then Google Deepmind, including Ian Osband, Geoffrey Irving, Rich Sutton, Shi Dong, Junhyuk Oh, Zeyu Zheng, Rishabh Joshi, David Silver, Satinder Baveja, Doina Precup, Yee Whye Teh, Jasper Snoek, James Harrison, Bo Dai, Kevin Murphy, and Shibl Mourad.

\bibliography{references}

@book{sutton2018reinforcement,
  title        = {Reinforcement Learning: An Introduction},
  author       = {Sutton, Richard S. and Barto, Andrew G.},
  edition      = {2},
  year         = {2018},
  publisher    = {MIT Press},
  address      = {Cambridge, MA},
}

@inproceedings{osband2023epistemic,
title={Epistemic Neural Networks},
author={Ian Osband and Zheng Wen and Seyed Mohammad Asghari and Vikranth Dwaracherla and Morteza Ibrahimi and Xiuyuan Lu and Van Roy, Benjamin},
booktitle={Thirty-seventh Conference on Neural Information Processing Systems},
year={2023}
}

@article{Loshchilov2017adamw,
  author       = {Ilya Loshchilov and
                  Frank Hutter},
  title        = {Fixing Weight Decay Regularization in {A}dam},
  journal      = {CoRR},
  volume       = {abs/1711.05101},
  year         = {2017},
  eprinttype    = {arXiv},
  eprint       = {1711.05101},
  bibsource    = {dblp computer science bibliography, https://dblp.org}
}

@InProceedings{dwaracherla2024efficient,
  title = 	 {Efficient Exploration for {LLM}s},
  author =       {Dwaracherla, Vikranth and Asghari, Seyed Mohammad and Hao, Botao and Van Roy, Benjamin},
  booktitle = 	 {Proceedings of the 41st International Conference on Machine Learning},
  pages = 	 {12215--12227},
  year = 	 {2024},
  editor = 	 {Salakhutdinov, Ruslan and Kolter, Zico and Heller, Katherine and Weller, Adrian and Oliver, Nuria and Scarlett, Jonathan and Berkenkamp, Felix},
  volume = 	 {235},
  series = 	 {Proceedings of Machine Learning Research},
  month = 	 {21--27 Jul},
  publisher =    {PMLR},
  url = 	 {https://proceedings.mlr.press/v235/dwaracherla24a.html}
}

@article{kaplan2020scaling,
  title={Scaling laws for neural language models},
  author={Kaplan, Jared and McCandlish, Sam and Henighan, Tom and Brown, Tom B and Chess, Benjamin and Child, Rewon and Gray, Scott and Radford, Alec and Wu, Jeffrey and Amodei, Dario},
  journal={arXiv preprint arXiv:2001.08361},
  year={2020}
}

@incollection{osband2018rpf,
	title = {Randomized Prior Functions for Deep Reinforcement Learning},
	author = {Osband, Ian and Aslanides, John and Cassirer, Albin},
	booktitle = {Advances in Neural Information Processing Systems 31},
	editor = {S. Bengio and H. Wallach and H. Larochelle and K. Grauman and N. Cesa-Bianchi and R. Garnett},
	pages = {8617--8629},
	year = {2018},
	publisher = {Curran Associates, Inc.}
}

@inproceedings{
abdolmaleki2025learningnegativefeedbackpositive,
title={Learning from negative feedback, or positive feedback or both},
author={Abbas Abdolmaleki and Bilal Piot and Bobak Shahriari and Jost Tobias Springenberg and Tim Hertweck and Michael Bloesch and Rishabh Joshi and Thomas Lampe and Junhyuk Oh and Nicolas Heess and Jonas Buchli and Martin Riedmiller},
booktitle={The Thirteenth International Conference on Learning Representations},
year={2025},
url={https://openreview.net/forum?id=4FVGowGzQb}
}

@article{irving2018ai,
  title={{AI} safety via debate},
  author={Irving, Geoffrey and Christiano, Paul and Amodei, Dario},
  journal={arXiv preprint arXiv:1805.00899},
  year={2018}
}

@article{marklund2024choice,
  title={Choice Between Partial Trajectories: Disentangling Goals from Beliefs},
  author={Marklund, Henrik and Van Roy, Benjamin},
  journal={arXiv preprint arXiv:2410.22690},
  year={2024}
}

@inproceedings{mehta2025sample,
title={Sample Efficient Preference Alignment in {LLM}s via Active Exploration},
author={Viraj Mehta and Syrine Belakaria and Vikramjeet Das and Ojash Neopane and Yijia Dai and Ilija Bogunovic and Barbara E Engelhardt and Stefano Ermon and Jeff Schneider and Willie Neiswanger},
booktitle={Second Conference on Language Modeling},
year={2025},
}

@article{dong2024rlhf,
    title={{RLHF} Workflow: From Reward Modeling to Online {RLHF}},
    author={Hanze Dong and Wei Xiong and Bo Pang and Haoxiang Wang and Han Zhao and Yingbo Zhou and Nan Jiang and Doyen Sahoo and Caiming Xiong and Tong Zhang},
    journal={Transactions on Machine Learning Research},
    issn={2835-8856},
    year={2024},
    note={}
}

@article{bai2022rlhf,
  title={Training a helpful and harmless assistant with reinforcement learning from human feedback},
  author={Bai, Yuntao and Jones, Andy and Ndousse, Kamal and Askell, Amanda and Chen, Anna and DasSarma, Nova and Drain, Dawn and Fort, Stanislav and Ganguli, Deep and Henighan, Tom and others},
  journal={arXiv preprint arXiv:2204.05862},
  year={2022}
}

@article{tang2024understanding,
  title={Understanding the performance gap between online and offline alignment algorithms},
  author={Tang, Yunhao and Guo, Daniel Zhaohan and Zheng, Zeyu and Calandriello, Daniele and Cao, Yuan and Tarassov, Eugene and Munos, R{\'e}mi and Pires, Bernardo {\'A}vila and Valko, Michal and Cheng, Yong and others},
  journal={arXiv preprint arXiv:2405.08448},
  year={2024}
}

@inproceedings{bose2025hybrid,
title={Hybrid Preference Optimization for Alignment: Provably Faster Convergence Rates by Combining Offline Preferences with Online Exploration},
author={Avinandan Bose and Zhihan Xiong and Aadirupa Saha and Simon Shaolei Du and Maryam Fazel},
booktitle={ICLR Workshop: Quantify Uncertainty and Hallucination in Foundation Models: The Next Frontier in Reliable AI},
year={2025},
}

@techreport{settles2009active,
  title={Active learning literature survey},
  author={Settles, Burr},
  year={2009},
  institution={University of Wisconsin-Madison Department of Computer Sciences}
}

@inproceedings{diao2023active,
  title={Active prompting with chain-of-thought for large language models},
  author={Diao, Shizhe and Wang, Pengcheng and Lin, Yong and Pan, Rui and Liu, Xiang and Zhang, Tong},
  booktitle={Proceedings of the 62nd Annual Meeting of the Association for Computational Linguistics (Volume 1: Long Papers)},
  pages={1330--1350},
  year={2024}
}

@article{bayer2026activellm,
    author = {Bayer, Markus and Lutz, Justin and Reuter, Christian},
    title = {ActiveLLM: Large Language Model-Based Active Learning for Textual Few-Shot Scenarios},
    journal = {Transactions of the Association for Computational Linguistics},
    volume = {14},
    pages = {1-22},
    year = {2026},
    month = {01},
    issn = {2307-387X},
    doi = {10.1162/TACL.a.63},
    url = {https://doi.org/10.1162/TACL.a.63},
    eprint = {https://direct.mit.edu/tacl/article-pdf/doi/10.1162/TACL.a.63/2574429/tacl.a.63.pdf},
}

@article{ji2025reinforcement,
title={Reinforcement Learning from Human Feedback with Active Queries},
author={Kaixuan Ji and Jiafan He and Quanquan Gu},
journal={Transactions on Machine Learning Research},
issn={2835-8856},
year={2025},
note={Featured Certification}
}

@inproceedings{das2025active,
  title={Active preference optimization for sample efficient {RLHF}},
  author={Das, Nirjhar and Chakraborty, Souradip and Pacchiano, Aldo and Chowdhury, Sayak Ray},
  booktitle={Joint European Conference on Machine Learning and Knowledge Discovery in Databases},
  pages={96--112},
  year={2025},
  organization={Springer}
}

@inproceedings{
lin2026activedpo,
title={Active{DPO}: Active Direct Preference Optimization for Sample-Efficient Alignment},
author={Xiaoqiang Lin and Arun Verma and Zhongxiang Dai and Daniela Rus and See-Kiong Ng and Bryan Kian Hsiang Low},
booktitle={The Fourteenth International Conference on Learning Representations},
year={2026},
url={https://openreview.net/forum?id=RD4XgyVyGh}
}

@InProceedings{pmlr-v235-muldrew24a,
  title = 	 {Active Preference Learning for Large Language Models},
  author =       {Muldrew, William and Hayes, Peter and Zhang, Mingtian and Barber, David},
  booktitle = 	 {Proceedings of the 41st International Conference on Machine Learning},
  pages = 	 {36577--36590},
  year = 	 {2024},
  editor = 	 {Salakhutdinov, Ruslan and Kolter, Zico and Heller, Katherine and Weller, Adrian and Oliver, Nuria and Scarlett, Jonathan and Berkenkamp, Felix},
  volume = 	 {235},
  series = 	 {Proceedings of Machine Learning Research},
  month = 	 {21--27 Jul},
  publisher =    {PMLR},
}

@inproceedings{xie2025exploratory,
title={Exploratory Preference Optimization: Harnessing Implicit {$Q^*$}-Approximation for Sample-Efficient {RLHF}},
author={Tengyang Xie and Dylan J Foster and Akshay Krishnamurthy and Corby Rosset and Ahmed Hassan Awadallah and Alexander Rakhlin},
booktitle={The Thirteenth International Conference on Learning Representations},
year={2025},
}

@ARTICLE{qi2025sample,
  author={Qi, Han and Yang, Haochen and Zhang, Qiaosheng and Yang, Zhuoran},
  journal={IEEE Transactions on Information Theory},
  title={Sample-Efficient Reinforcement Learning From Human Feedback via Information-Directed Sampling},
  year={2025},
  volume={71},
  number={10},
  pages={7942-7958}}

@inproceedings{
belakaria2025sharpe,
title={Sharpe Ratio-Guided Active Learning for Preference Optimization in {RLHF}},
author={Syrine Belakaria and Joshua Kazdan and Charles Marx and Chris Cundy and Willie Neiswanger and Sanmi Koyejo and Barbara E Engelhardt and Stefano Ermon},
booktitle={Second Conference on Language Modeling},
year={2025},
url={https://openreview.net/forum?id=a6xzTqMUFQ}
}

@article{liu2024sample,
  title={Sample-efficient alignment for {LLM}s},
  author={Liu, Zichen and Chen, Changyu and Du, Chao and Lee, Wee Sun and Lin, Min},
  journal={arXiv preprint arXiv:2411.01493},
  year={2024}
}

@inproceedings{hoffmann2022scalinglaws,
author = {Hoffmann, Jordan and Borgeaud, Sebastian and Mensch, Arthur and Buchatskaya, Elena and Cai, Trevor and Rutherford, Eliza and de Las Casas, Diego and Hendricks, Lisa Anne and Welbl, Johannes and Clark, Aidan and Hennigan, Tom and Noland, Eric and Millican, Katie and van den Driessche, George and Damoc, Bogdan and Guy, Aurelia and Osindero, Simon and Simonyan, Karen and Elsen, Erich and Vinyals, Oriol and Rae, Jack W. and Sifre, Laurent},
title = {Training compute-optimal large language models},
year = {2022},
isbn = {9781713871088},
publisher = {Curran Associates Inc.},
address = {Red Hook, NY, USA},
booktitle = {Proceedings of the 36th International Conference on Neural Information Processing Systems},
articleno = {2176},
numpages = {15},
location = {New Orleans, LA, USA},
series = {NIPS '22}
}

@misc{yuan2023scaling-sft,
      title={Scaling Relationship on Learning Mathematical Reasoning with Large Language Models},
      author={Zheng Yuan and Hongyi Yuan and Chengpeng Li and Guanting Dong and Keming Lu and Chuanqi Tan and Chang Zhou and Jingren Zhou},
      year={2023},
      eprint={2308.01825},
      archivePrefix={arXiv},
      primaryClass={cs.CL},
}

@misc{gao2022scalinglawsrewardmodel,
      title={Scaling Laws for Reward Model Overoptimization},
      author={Leo Gao and John Schulman and Jacob Hilton},
      year={2022},
      eprint={2210.10760},
      archivePrefix={arXiv},
      primaryClass={cs.LG},
}

@inproceedings{rafailov2024scaling,
author = {Rafailov, Rafael and Chittepu, Yaswanth and Park, Ryan and Sikchi, Harshit and Hejna, Joey and Knox, W. Bradley and Finn, Chelsea and Niekum, Scott},
title = {Scaling laws for reward model overoptimization in direct alignment algorithms},
year = {2025},
isbn = {9798331314385},
publisher = {Curran Associates Inc.},
address = {Red Hook, NY, USA},
booktitle = {Proceedings of the 38th International Conference on Neural Information Processing Systems},
articleno = {4009},
numpages = {36},
location = {Vancouver, BC, Canada},
series = {NIPS '24}
}

@InProceedings{xiong2024iterativepreferencelearninghuman,
  title = 	 {Iterative Preference Learning from Human Feedback: Bridging Theory and Practice for {RLHF} under {KL}-constraint},
  author =       {Xiong, Wei and Dong, Hanze and Ye, Chenlu and Wang, Ziqi and Zhong, Han and Ji, Heng and Jiang, Nan and Zhang, Tong},
  booktitle = 	 {Proceedings of the 41st International Conference on Machine Learning},
  pages = 	 {54715--54754},
  year = 	 {2024},
  editor = 	 {Salakhutdinov, Ruslan and Kolter, Zico and Heller, Katherine and Weller, Adrian and Oliver, Nuria and Scarlett, Jonathan and Berkenkamp, Felix},
  volume = 	 {235},
  series = 	 {Proceedings of Machine Learning Research},
  month = 	 {21--27 Jul},
  publisher =    {PMLR}
}

@misc{hou2024doesrlhfscaleexploring,
      title={Does {RLHF} Scale? Exploring the Impacts From Data, Model, and Method},
      author={Zhenyu Hou and Pengfan Du and Yilin Niu and Zhengxiao Du and Aohan Zeng and Xiao Liu and Minlie Huang and Hongning Wang and Jie Tang and Yuxiao Dong},
      year={2024},
      eprint={2412.06000},
      archivePrefix={arXiv},
      primaryClass={cs.CL},
}

@misc{cobbe2021trainingverifierssolvemath,
      title={Training Verifiers to Solve Math Word Problems},
      author={Karl Cobbe and Vineet Kosaraju and Mohammad Bavarian and Mark Chen and Heewoo Jun and Lukasz Kaiser and Matthias Plappert and Jerry Tworek and Jacob Hilton and Reiichiro Nakano and Christopher Hesse and John Schulman},
      year={2021},
      eprint={2110.14168},
      archivePrefix={arXiv},
      primaryClass={cs.LG},
}

@misc{guo2024directlanguagemodelalignment,
      title={Direct Language Model Alignment from Online {AI} Feedback},
      author={Shangmin Guo and Biao Zhang and Tianlin Liu and Tianqi Liu and Misha Khalman and Felipe Llinares and Alexandre Rame and Thomas Mesnard and Yao Zhao and Bilal Piot and Johan Ferret and Mathieu Blondel},
      year={2024},
      eprint={2402.04792},
      archivePrefix={arXiv},
      primaryClass={cs.AI},
}

@misc{geminiteam2024gemini15unlockingmultimodal,
      title={Gemini 1.5: Unlocking multimodal understanding across millions of tokens of context},
      author={{Gemini Team}},
      year={2024},
      eprint={2403.05530},
      archivePrefix={arXiv},
      primaryClass={cs.CL},
      url={https://arxiv.org/abs/2403.05530},
}

@misc{gemmateam2024gemma2improvingopen,
      title={Gemma 2: Improving Open Language Models at a Practical Size},
      author={Gemma Team and Morgane Riviere and Shreya Pathak and Pier Giuseppe Sessa and Cassidy Hardin and Surya Bhupatiraju and Léonard Hussenot and Thomas Mesnard and Bobak Shahriari and others},
      year={2024},
      eprint={2408.00118},
      archivePrefix={arXiv},
      primaryClass={cs.CL},
      url={https://arxiv.org/abs/2408.00118},
}

@article{Bradley1952Rank,
 author = {Ralph Allan Bradley and Milton E. Terry},
 journal = {Biometrika},
 number = {3/4},
 pages = {324--345},
 publisher = {[Oxford University Press, Biometrika Trust]},
 title = {Rank Analysis of Incomplete Block Designs: I. The Method of Paired Comparisons},
 urldate = {2025-11-24},
 volume = {39},
 year = {1952}
}

\appendix

\end{document}